%% file: Melodi-arxiv-v1.tex
\documentclass{article} 
\usepackage{iclr2025_conference,times}

\input{math_commands.tex}

\usepackage{hyperref}
\usepackage{url}

\usepackage{graphicx}
\usepackage{subfigure}
\usepackage{wrapfig}
\usepackage{bm}
\usepackage{caption}
\usepackage{multirow}

\usepackage{ulem}

\title{Melodi: Exploring Memory Compression for Long Contexts}

\author{\quad\quad Yinpeng Chen, \quad\quad DeLesley Hutchins, \quad\quad Aren Jansen, \quad\quad Andrey Zhmoginov, \AND \quad\quad\quad\quad\quad\quad\quad\quad\quad\quad\quad David Racz, \quad\quad  Jesper Andersen \\ \\[2mm]
\quad\quad\quad\quad\quad\quad\quad\quad\quad\quad\quad\quad\quad\quad Google DeepMind \\[1mm]
\texttt{\{yinpengc,delesley,arenjansen,azhmogin,dracz,jespera\}@google.com} \\
}

\def\melodi{\textsc{Melodi}}

\iclrfinalcopy 
\begin{document}

\maketitle

\begin{abstract}
We present \melodi, a novel memory architecture designed to efficiently process long documents using short context windows. The key principle behind \melodi{} is to represent short-term and long-term memory as a hierarchical compression scheme across both network layers and context windows. Specifically, the short-term memory is achieved through recurrent compression of context windows across multiple layers, ensuring smooth transitions between windows. In contrast, the long-term memory performs further compression within a single middle layer and aggregates information across context windows, effectively consolidating crucial information from the entire history. Compared to a strong baseline - the Memorizing Transformer employing dense attention over a large long-term memory (64K key-value pairs) - our method demonstrates superior performance on various long-context datasets while remarkably reducing the memory footprint by a factor of 8.
\end{abstract}

\section{Introduction}
\vspace{-4mm}
\begin{figure}[htb!]
\begin{center}
\centerline{\includegraphics[width=0.8\columnwidth]{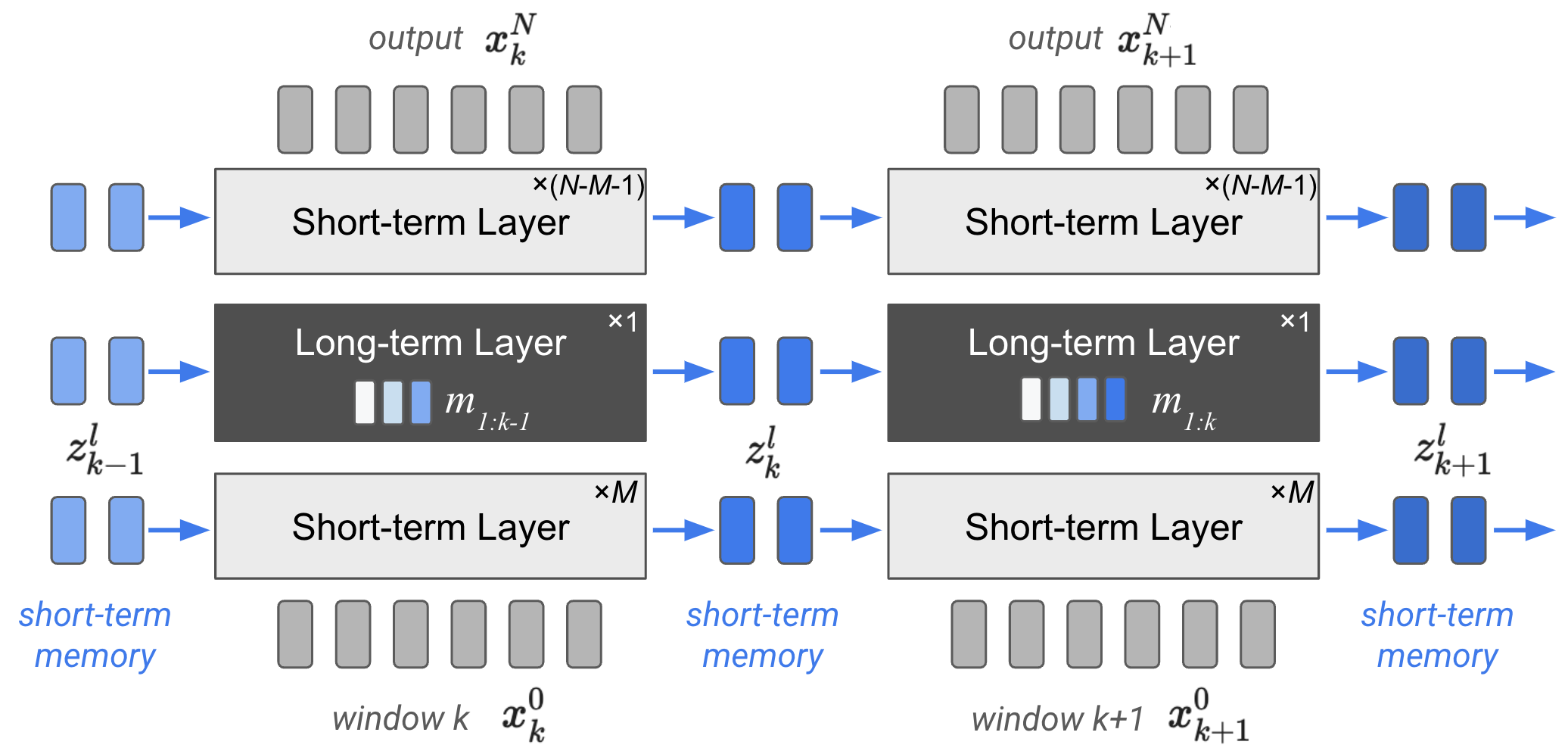}}
\vspace{-2mm}
\caption{\textbf{Overview of \melodi.} \melodi{} employs a hierarchical memory representation, incorporating both short-term and long-term compression mechanisms, integrated with a transformer-based language model. It utilizes a stack of short-term layers to recurrently compress each context window $x_k^0$ into short-term memory tokens $\{z^l_k\}$, and inserts a long-term layer to store compressed key-value pairs within a long-term memory $m_{1:k}$. Both short-term and long-term layers leverage modified transformer blocks. In this illustration, we assume a total of $N$ layers, with $M$ short-term layers preceding 1 long-term layer and $N-M-1$ short-term layers following it.}
\label{fig:melodi-intro}
\end{center}
\vspace{-6mm}
\end{figure}

Long-context language models, exemplified by Gemini \citep{geminiteam2024gemini15unlockingmultimodal} and GPT \citep{openai2024gpt4technicalreport}, showcase remarkable capabilities across diverse modalities (e.g., text, images, audio, code, video) and seamlessly integrate various machine learning techniques, including many-shot in-context learning \citep{agarwal2024manyshotincontextlearning}, chain-of-thought prompting \citep{NEURIPS2022CoT}, and the incorporation of explicit instructions \citep{JMLR:v25:scaling-instruction-finetuned-LM, wei2022finetuned}. However, the quadratic complexity of attention mechanisms within transformer models necessitates significant computational resources to handle long contexts effectively.  This has spurred the development of efficient solutions \citep{dai-etal-2019-transformer, wu2022memorizing, NEURIPS2022_rmt} that process long contexts via short context windows, much like how humans process information by reading a book chapter by chapter.  A central question underlying these solutions is: {\it how can we effectively model and manage memory to bridge the gaps between these short context windows over long context?}

Memory fundamentally revolves around compressing and storing information for future utilization, all within the constraints of limited capacity. The Long Short-Term Memory (LSTM) architecture \citep{HochSchm97} tackles this by recurrently compressing historical information into hidden states after processing each token. With the rise of Transformer models \citep{NIPS2017_transformer} dominating the language modeling landscape, recent memory designs have shifted towards utilizing Transformers to process a context window, thereby moving the focus of memory management from the token level to the context window level.

Transformer-XL \citep{dai-etal-2019-transformer} employs a caching mechanism to store multi-layer key-value (KV) pairs from the preceding window as memory. Memorizing Transformer \citep{wu2022memorizing} builds upon this foundation by incorporating a dedicated layer to memorize all KV pairs from that layer across all prior windows.  Meanwhile, Block Recurrent Transformer \citep{NEURIPS2022_block_recurrent} and Recurrent Memory Transformer \citep{NEURIPS2022_rmt} introduce distinct recurrent compression mechanisms, implemented in a middle layer and at the output, respectively.

In this paper, we introduce \melodi{} (short for ``\underline{ME}mory with \underline{LO}w \underline{DI}mension''), an efficient memory architecture designed to handle long contexts despite operating on short context windows (e.g., 512 tokens per window). \melodi{} integrates both short-term and long-term memory through a compression-based approach. The short-term memory, recurrent in nature and possessing low capacity, spans multiple network layers, progressively compressing context tokens and prior memory at each layer. In contrast, the long-term memory, incremental and high-capacity, resides within a single network layer. It maintains a record of the entire history by further compressing each context window and stacking them.  Both short-term and long-term memory are seamlessly incorporated into a multi-layer transformer model using a ``sandwich'' structure (see Figure~\ref{fig:melodi-intro}), incurring negligible additional parameters.

\melodi{} demonstrates strong performance on various long-context datasets. For instance, utilizing a 13-layer transformer network with 1024 embedding dimensions and 512-token context windows, \melodi{} achieves perplexity scores of 10.44 and 2.11 on PG-19 (T5 vocabulary) and arXiv Math (Meena vocabulary), respectively.  This represents a clear improvement over the Memorizing Transformer (10.62 on PG-19, 2.14 on arXiv) with dense attention (as opposed to top-k attention), while significantly reducing memory usage by a factor of 8.  Furthermore, ablation studies confirm the complementary nature of short-term and long-term memory in \melodi{}, highlighting their synergistic contribution to an efficient and effective memory architecture.

\section{Our Method: Melodi}
\label{sec:melodi}

\melodi{} focuses on efficiently comprehending \textit{long contexts} through the utilization of \textit{short context windows}, thus circumventing the quadratic complexity associated with attention mechanisms over long sequences. This approach necessitates a memory design that not only ensures smooth transitions between windows but also preserves crucial information from all preceding windows.


\subsection{Architecture Overview}
\textbf{Design principle:}
The core principle behind \melodi{} is to represent short-term and long-term memory through a hierarchical compression scheme. Specifically, the short-term memory recurrently compresses context tokens across multiple network layers (e.g., condensing a 512-token context window into 128 memory tokens). This process not only facilitates seamless transitions between context windows but also aggregates information across them, effectively functioning as a fixed-size multi-layer long short-term memory (LSTM) mechanism \citep{HochSchm97}. Furthermore, each context window undergoes additional compression within a middle layer and is then stacked into a long-term memory.  This long-term memory retains essential information from the entire history, thus compensating for any potential forgetting in the short-term memory. Both short-term and long-term memory are seamlessly integrated into a transformer-based language model, enabling the comprehension of long contexts even under the constraint of short context windows.

\textbf{Sandwich architecture:}
\melodi{}'s network architecture (Figure~\ref{fig:melodi-intro}) features a "sandwich" structure, with a long-term compression layer inserted between multiple recurrent short-term compression layers. Both layer types utilize a standard transformer block (attention and feed-forward network) with tailored compression modifications. The short-term layers recurrently compress the current context window and update short-term memory, while the long-term layer further compresses information and appends it to long-term memory.

\textbf{Terminology:}
In the remainder of this paper, we adopt the following notation.  We use  $k$ to index context windows and $l$ to index network layers. Within the  $l^{th}$ layer of the $k^{th}$ context window, the input context tokens are represented by $x_k^{l-1}$.  The output context tokens, denoted as $x_k^l$, serve as input for the subsequent layer. The input and output of the short-term memory are $z_{k-1}^l$ and $z_{k}^l$, respectively. The long-term memory preceding window $k$ is denoted as $m_{1:k-1}$. Note that we omit the subscript $l$ for the long-term memory since it resides within a single long-term layer.

Next, we will discuss both short-term and long-term layers in detail. 

\subsection{Short-Term Memory: Multi-Layer Recurrent Compression}

\begin{wrapfigure}{r}{0.47\textwidth}
\vspace{-6mm}
\begin{center}
\centerline{\includegraphics[width=0.47\columnwidth]{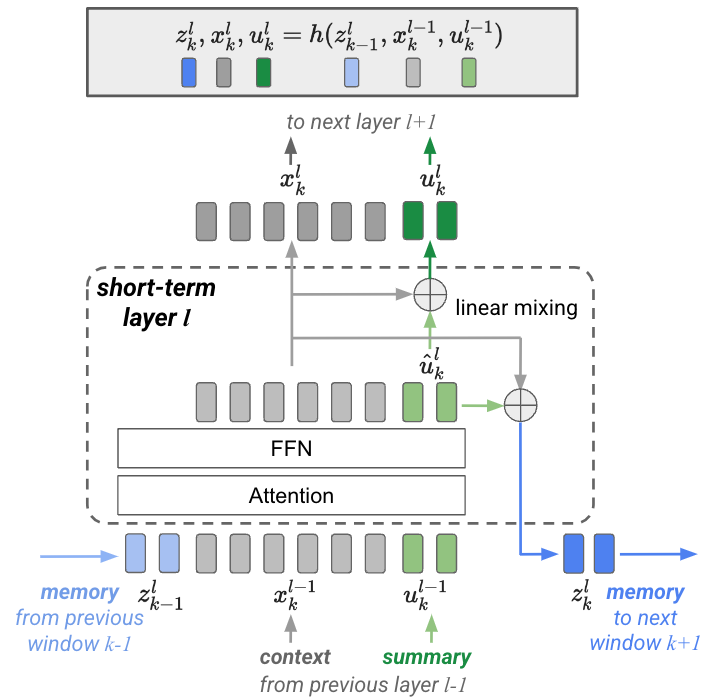}}
\vspace{-2mm}
\caption{\textbf{Short-term layer.} The figure illustrates the processing of the $k^{th}$ context window at the $l^{th}$ short-term layer. It takes the memory from the previous window $z_{k-1}^l$ and the current context/summary ($x_{k}^{l-1}$, $u_{k}^{l-1}$) from the previous layer as input. The short-term layer adds two linear token mixers \citep{tolstikhin2021mlpmixer} on top of a standard transformer layer (including attention and FFN) to separate the summary for the next layer $u_{k}^{l}$ and the memory for the next window $z_{k}^l$. Best viewed in color.}
\label{fig:short-term}
\end{center}
\vspace{-10mm}
\end{wrapfigure}

The short-term memory is distributed across multiple short-term layers (see Figure~\ref{fig:melodi-intro}). This subsection delves into the specifics of the short-term layer, using the $l^{th}$ layer as an illustrative example for processing the $k^{th}$ context window.
The short-term layer serves a dual purpose: (a) transforming context tokens $x_k^{l-1}$ via a transformer block (yielding output $x_k^l$), and (b) recurrently compressing the current context window into the short-term memory $z_k^l$. It accomplishes this by updating both context tokens and short-term memory through a shared transformer block, albeit along separate pathways. As visualized in Figure~\ref{fig:short-term}, context tokens traverse vertically across layers (from $x_k^{l-1}$ to $x_k^l$), whereas short-term memory flows horizontally across context windows (from $z_{k-1}^l$ to $z_k^l$). To enable inter-layer communication within the short-term memory, we introduce summary tokens $u_k^l$ that propagate through the layers. We elaborate on the key components below.

\textbf{Short-term memory $z_k^l$:} 
The short-term memory (illustrated in Figure~\ref{fig:short-term}) is implemented as a sequence of vectors with length $S$, each having the same dimensionality as context tokens (e.g., 1024 channels). Notably, the number of short-term memory vectors is substantially smaller than the length of the context window (e.g., 128 memory tokens per window of 512 context tokens). Within each context window, the short-term memory serves initially as a previous context for auto-regressive prediction of subsequent context tokens (except for the first window in which the short-term memory is empty). It is then updated by compressing and incorporating information from the context tokens within the current window.

\textbf{Transforming context tokens:}
The context tokens $x_k^l$ are generated through causal attention to both (a) the preceding short-term memory $z_{k-1}^l$ and (b) preceding tokens within the current context window. This attention mechanism is followed by the application of a feed-forward network (FFN). Relative position embeddings are employed for both the context tokens $x_k^{l-1}$ and the preceding short-term memory $z_{k-1}^l$. Mathematically, this can be represented as:
\begin{align}
 x_k^l=\mathcal{T}(x_k^{l-1}|z_{k-1}^l),
 \label{eq:transformer-x}
\end{align}    
where $\mathcal{T}(x|z)$ indicates applying a transformer block on $x$ for a given preceding context $z$.

\textbf{Recurrent compression:}
Beyond transforming context tokens, the short-term layer also recurrently compresses the current context window into short-term memory. Similar to the approach in RMT \citep{NEURIPS2022_rmt} and AutoCompressors \citep{chevalier-etal-2023-auto-compressor}, this compression is achieved by appending summary tokens $u$ after context tokens $x$ and passing the combined sequence through the transformer block. Consequently, the resulting summary tokens compresses both the preceding short-term memory and the current context window via attentional pooling, expressed as: $\hat{u}_k^l=\mathcal{T}(u_k^{l-1}|z_{k-1}^l, x_k^{l-1})$, where the input summary tokens $u_k^{l-1}$ originate from the previous layer (refer to Figure~\ref{fig:short-term}). We use $\hat{u}_k^l$ (instead of $u_k^l$) to denote an intermediate result that is further processed in the subsequent summary branching step. In practice, both context and summary tokens can be updated simultaneously within a single transformer operation: $x_k^l, \hat{u}_k^l =\mathcal{T}(x_k^{l-1}, u_k^{l-1}|z_{k-1}^l)$. Relative position embeddings are applied on the short-term memory $z_{k-1}^l$, context $x_{k}^{l-1}$, and summary $u_{k}^{l-1}$, while a causal mask is applied on the combined sequence of $x_{k}^{l-1}$ and $u_{k}^{l-1}$. 

Summary tokens (containing $U$ tokens) are initialized from learnable embeddings (prior to the first layer) and set to the same length as the short-term memory ($U=S$).  Propagating through all layers, they facilitate inter-layer communication within the short-term memory.
%
Moreover, branching summary tokens both upwards to the next layer and rightwards to the next window improves performance, a strategy we will discuss in more detail subsequently.

\textbf{Summary branching:} 
We employ distinct linear token mixers \citep{tolstikhin2021mlpmixer} on the context and summary tokens to generate separate summary tokens for the subsequent layer and short-term memory tokens for the next window. Unlike channel mixing, 
a linear token mixer linearly combines the $M_i$ input tokens across each channel to produce $M_o$ output tokens with the same dimensionality by using an $M_i\times M_o$ matrix. This is mathematically represented as follows:
%
%
\begin{align}
 u_k^l=\mathcal{M}_{\uparrow}(x_k^l, \hat{u}_k^l), \;\;
 z_k^l=\mathcal{M}_{\rightarrow}(x_k^l, \hat{u}_k^l), \;\;
 \text{where}\;\;
 x_k^l, \hat{u}_k^l=\mathcal{T}(x_k^{l-1}, u_k^{l-1}|z_{k-1}^l),
 \label{eq:finola-intro}
\end{align}    
where $\mathcal{M}_{\uparrow}$ and $\mathcal{M}_{\rightarrow}$ denote the two linear token mixers towards the subsequent layer and the next window respectively. The resulting summary and memory tokens exhibit distinct linear combination of context $x_k^l$ and summary tokens $\hat{u}_k^l$, implying divergent compression flows across layers and windows. Since the summary and short-term memory share the same number of tokens ($S$), each mixer
 comprises $(W+S)\times S$ parameters, where $W$ represent the number of context tokens per window. This parameter count is negligible for short context windows. For instance, with a context window of 512 tokens and 128 summary tokens, the two mixers collectively require (512+128)$\times$128$\times$2=164K parameters, constituting a mere 1.3\% of a transformer block with 1024 dimensions.

\textbf{Summary:} 
The short-term memory layer can be succinctly represented as a function $z_k^l, x_k^l, u_k^l=h(z_{k-1}^l, x_k^{l-1}, u_k^{l-1})$. This function transforms context tokens $x_k^{l-1}$ and summary tokens $u_k^{l-1}$ upward to the next layer (from $l-1$ to $l$) while simultaneously propagating short-term memory $z_{k-1}^l$ rightward to the next window (from $k-1$ to  $k$). Built upon a standard transformer block, this layer introduces negligible additional parameters through the summary branching mechanism.


\textbf{Relation to Block Recurrent Transformer (BRT) \citep{NEURIPS2022_block_recurrent}:} 
While BRT represents short-term memory by combining multi-layer KV caching from Transformer XL with a dedicated recurrent memory layer, Melodi models short-term memory as a consistent recurrent compression mechanism across multiple layers. Moreover, \melodi{} updates the short-term tokens by adding residual connections (cross-attending to the previous memory) over the previous layer, instead of over the previous memory tokens directly as in BRT.

While short-term memory facilitates smooth transitions between context windows, the inherent limitation of its capacity can lead to the inevitable loss of information, particularly for contexts located further back in the sequence.  In the following subsection, we will demonstrate how long-term memory can be leveraged to mitigate this forgetting issue.

\subsection{Long-Term Memory: Single-Layer Memorizing Compressed Key-Value Pairs}

\begin{figure}[t]
\begin{center}
\centerline{\includegraphics[width=0.9\columnwidth]{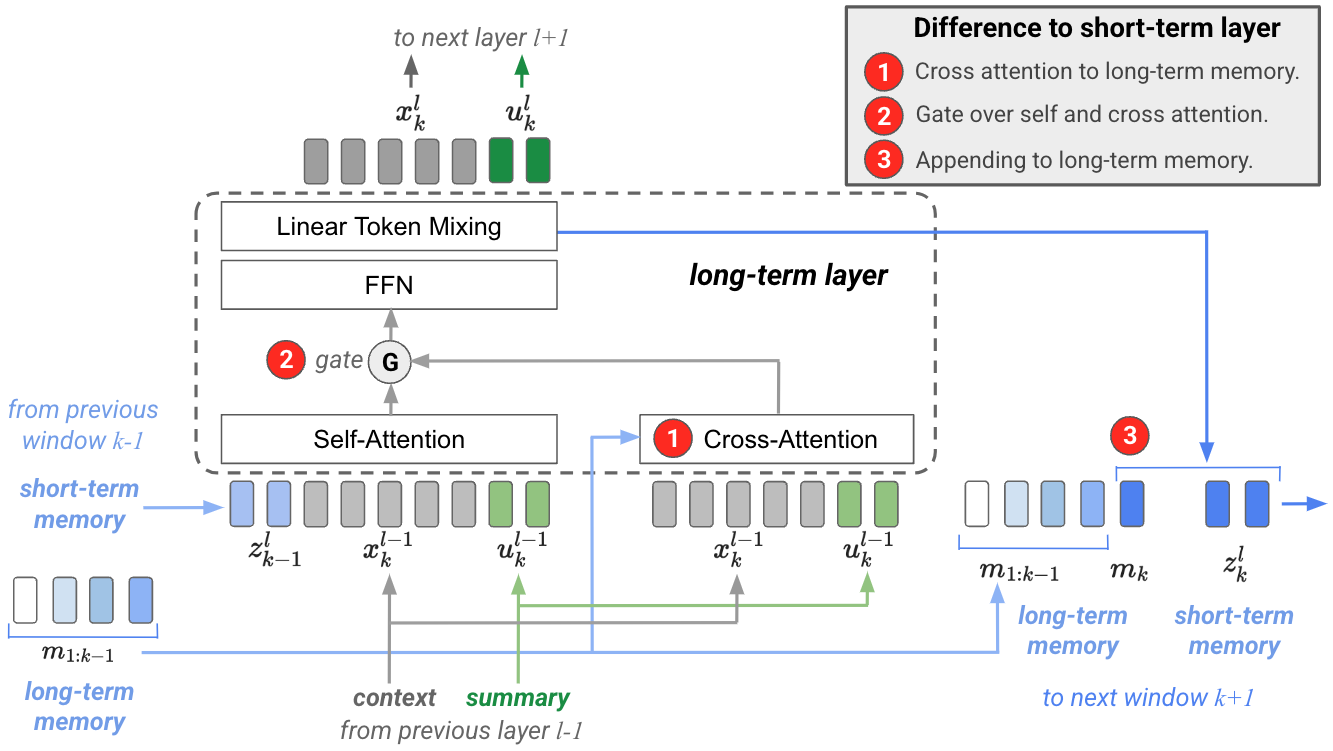}}
\vspace{-2mm}
\caption{\textbf{Long-term layer.} The long-term layer adds three components on top of the short-term layer (see Figure~\ref{fig:short-term}). Firstly, it introduces a long-term memory $m_{1:k-1}$ by caching the compressed key-value pairs and allows the current context/summary ($x_{k}^{l-1}$, $u_{k}^{l-1}$) to cross attend to them. Secondly, the self-attention and cross-attention are integrated via gating. Finally, the linear token mixing output additional compressed tokens and appends their key-value pairs $m_k$ into the long-term memory (as $m_{1:k}$) for the next window. Best viewed in color.}
\label{fig:long-term}
\end{center}
\vspace{-10mm}
\end{figure}

In this subsection, we utilize long-term memory to retain information from all previous windows, thereby alleviating the forgetting inherent in short-term memory. The key idea involves further compressing the context window and storing the compressed representations across the entire history.

\textbf{Further compression:}
In contrast to the short-term memory, which compresses information across multiple layers, the long-term memory achieves a higher compression rate within \textit{a single layer}.  Specifically, it compresses a context window into $L$ long-term tokens at a designated middle layer (refer to Figure~\ref{fig:melodi-intro}), where $L$ is less than the number of short-term tokens per layer ($L<S$).  Illustratively, we might compress a context window of 512 tokens into $S=128$ short-term tokens at each layer, but further condense it into $L=64$ long-term tokens at a single layer.

\textbf{Storing long-term memory:}
The key-value (KV) pairs of long-term tokens are sequentially stored in a first-in-first-out (FIFO) queue with a maximum capacity of $Q_{max}$ windows. Consequently, the long-term memory can hold up to $L\times Q_{max}$ KV pairs. For contexts shorter than $Q_{max}$ windows, a compressed representation of the entire prior history is preserved.  For longer documents exceeding $Q_{max}$ windows, a substantial portion of the recent history ($Q_{max}$ windows) is still retained. We opt to store KV pairs (rather than the tokens themselves) because they are repeatedly utilized in cross-attention mechanisms for subsequent context windows, a point we will elaborate on later.

\textbf{Long-term layer:}
Figure~\ref{fig:long-term} illustrates the implementation of a long-term layer, which builds upon a short-term layer but incorporates three key additions. First, it introduces a long-term memory component (denoted as $m_{1:k-1}$ prior to the $k^{th}$ context window) and enables the current context tokens $x_k$ and summary tokens $u_k$ to cross-attend to it. Second, the cross-attention shares parameters with the self-attention and their results (cross attention: $\mathcal{A}_x$, self-attention: $\mathcal{A}_s$) are combined through a gating mechanism using a learnable scalar $\alpha$ per attention head, formulated as $\alpha\mathcal{A}_x+(1-\alpha)\mathcal{A}_s$. Finally, an additional linear token mixer is introduced to generate long-term tokens for the current window, and their key-value (KV) pairs $m_k$ are appended to the long-term memory. This token mixer comprises $(W+U)\times L$ parameters, where $W$, $U$ and $L$ represent the number of context, summary and long-term memory tokens, respectively. It is notably smaller than the mixers in the short-term layer because: (a) $L$ is less than $S$ (the number of short-term memory tokens), and (b) it is present in only one layer.

\textbf{Long-term vs short-term memory:} 
Table~\ref{table:short-vs-long} provides a comparative overview of the long-term and short-term memory mechanisms. In contrast to the short-term memory, which operates recurrently across multiple layers, the long-term memory functions incrementally within a single layer. The short-term memory spans $N$ network layers, with each layer utilizing $S$ tokens, while the long-term memory queue spans $Q_{max}$ context windows, with each window represented by $L$ tokens.
To illustrate, consider a 12-layer transformer model processing context windows of 512 tokens each, with embedding dimension 1024.  If we employ $S=128$ short-term tokens per layer and $L=64$ long-term tokens per window, with a maximum capacity of $Q_{max}=128$ windows for the long-term memory, the short-term and long-term memory caches would have capacities of 1.6M (128$^{tokens}\times$1024 $^{dim}\times$12 $^{layers}$) and 16.8M (64$^{pairs}\times$2$^{tokens/pair}\times$1024$^{dim}\times$128$^{windows}$) floats, respectively. It's worth noting that, for computational efficiency, we store key-value (KV) pairs in the long-term memory, whereas tokens are stored directly in the short-term memory.

\textbf{Relation to Memorizing Transformer (MT) \citep{wu2022memorizing}:}
Similar to Memorizing Transformer, \melodi{} incorporates key-value pairs from a middle layer into its long-term memory.  Experiments corroborate the finding in MT that employing additional long-term layers yields only incremental gains.  However, a key distinction lies in the fact that \melodi{} stores \textit{compressed} KV pairs, rather than the KV pairs of context tokens directly, as in MT. This modification substantially reduces the size of the long-term memory.  For instance, when compressing a context window of 512 tokens into 64 long-term tokens, \melodi{} achieves an \textit{8-fold reduction} in long-term memory size.

\section{Experimental Results}
We evaluate \melodi{} on three long-context datasets: PG19 \citep{raecompressive2019}, arXiv Math \citep{wu2022memorizing}, and C4 \citep{2020c4}, using the standard auto-regressive language modeling task, where the objective is to predict the next token in a sequence. All models are trained from scratch, and we report the average perplexity on the respective test sets as our evaluation metric.

\subsection{Datasets}
\textbf{PG19:}
The PG19 dataset \citep{raecompressive2019} consists of 28,752 English books published before 1919, averaging around 68,972 tokens per book. We utilize three 32k vocabularies: (a) Meena \citep{thoppilan2022lamdalanguagemodelsdialog}, (b) T5 \citep{JMLR:v21:T5} and (c) a custom SentencePiece vocabulary \citep{kudo-richardson-2018-sentencepiece} trained specifically on PG19.

\textbf{arXiv Math:}
The arXiv dataset \citep{wu2022memorizing} comprises technical math papers from arXiv, with token counts comparable to PG19 due to LaTeX's use of special characters, resulting in smaller subwords. We use a 32k Meena vocabulary \citep{thoppilan2022lamdalanguagemodelsdialog} and a 32k custom vocabulary.



\begin{table*}[t!]
\begin{minipage}[t]{0.465\linewidth}
        \caption{\textbf{Long-term vs. short-term memory.} In our notation,  $L$ and $S$  represent the number of long-term and short-term tokens per network layer, respectively. Note that long-term has fewer tokens ($L<S$).  $Q_{max}$ denotes the maximum number of windows encompassed by the long-term memory queue.  $N$  signifies the total number of network layers.}
\label{table:short-vs-long}
\vspace{-4mm}
\begin{center}
\setlength{\tabcolsep}{1.0mm}{
\begin{tabular}{lcc}
\multicolumn{1}{c}{\bf PROPERTY}  &\multicolumn{1}{c}{\bf LONG} & \multicolumn{1}{c}{\bf SHORT}
\\ \hline \\[-2mm]
{\bf Number of layers} & \textit{single} & {\it multiple} \\
{\bf Update per window}  & {\it incremental} & {\it recurrent} \\
{\bf Capacity}    & $L\times Q_{max}$ & $S \times N$\\
\end{tabular}
}
\end{center}
\end{minipage}
\quad
    \begin{minipage}[t]{0.505\linewidth}
	\caption{\textbf{Baseline implementations.} Our re-implementations utilize a cosine decay learning rate schedule (replacing inverse square root decay) and dense cross-attention for the Memorizing Transformer (replacing top-k attention). This results in improved performance compared to prior reported results.}
\label{table:baseline}
\vspace{-3mm}
\begin{center}
\setlength{\tabcolsep}{0.8mm}{
\begin{tabular}{l|cc|cc}
\multicolumn{1}{c|}{\bf Baseline}  &\multicolumn{2}{c|}{{\bf PG-19} \small{(T5)}} & \multicolumn{2}{c}{{\bf arXiv} \small{(Meena)}} \\
\multicolumn{1}{c|}{\bf Method}  & prior & {\bf our} & prior & {\bf our} \\
 \hline 
& & & &\\[-2mm]
Transformer XL & 11.96 & {\bf 11.54} & 2.67 & {\bf 2.61} \\
Block Recurrent  & 11.55 & {\bf 10.98} & 2.36 & {\bf 2.26} \\
Memorizing Trans.    & 11.62 & {\bf 10.74} & 2.31 & {\bf 2.15} \\
\end{tabular}
}
\end{center}
    \end{minipage}  
    \vspace{-3mm}
\end{table*}

\textbf{C4(4K+):}
The C4 dataset \citep{2020c4} is a large collection of internet documents. To emphasize long documents where memory is crucial, we filter out those with fewer than 4,096 tokens and utilize a 32k custom vocabulary.


\subsection{Setup}
%
We utilized a decoder-only transformer architecture (with 12 or 13 layers), incorporating both short-term and long-term memory caches. The model had an embedding size of 1024, 8 attention heads with a dimensionality of 128 each, and an FFN hidden layer of size 4096. All models were implemented in JAX and Flax and trained from scratch for 500k steps on 32 TPU cores. Further training details are provided in the Appendix~\ref{appendix:training-details}.


During training, each long document was segmented into 4096-token chunks to facilitate batch processing. These chunks were then organized into training batches, each comprising 8 context windows of 512 tokens. In our ablation study, \melodi{}'s default configuration compressed each context window into $S$=128 short-term memory tokens per layer and $L$=64 long-term memory tokens.

\subsection{Comparison with Baselines}

\textbf{Baselines:} 
We benchmark \melodi{} against three well-established prior works: Transformer XL \citep{dai-etal-2019-transformer}, Block Recurrent Transformer \citep{NEURIPS2022_block_recurrent}, and Memorizing Transformer \citep{wu2022memorizing}. To ensure a fair comparison, we re-implement these baselines within our framework and evaluate them under identical settings.

Our re-implementations of the baseline models achieve superior performance compared to the results reported in their original papers (see Table \ref{table:baseline}). This improvement can be primarily attributed to two key factors: (a) using cosine decay learning rate schedule \citep{hoffmann2022trainingcomputeoptimallargelanguage} instead of the inverse square root decay, (b) using dense cross-attention instead of top-k attention for the Memorizing Transformer. Our re-implementations provide stronger baselines against which to evaluate \melodi{}'s effectiveness.

\begin{table}[t]
\caption{\textbf{Comparisons with baselines on three datasets}. The table reports average token-level perplexities for various models on three datasets. All models were trained under the same settings (e.g.  segment length 4096, context window 512, 500k training steps). Three \melodi{} configurations were used: $S_{192}+L_{32}$, $S_{128}+L_{64}$, and $S_{192}+L_{96}$. For instance, $S_{192}+L_{32}$ indicates $S=192$ short-term and $L=32$ long-term tokens per context window. All \melodi{} models utilized a long-term memory spanning 128 context windows.
\melodi{} $S_{192}+L_{32}$ outperformed Transformer XL and Block Recurrent Transformer while consuming less memory. Notably, \melodi{} $S_{192}+L_{96}$ clearly surpassed Memorizing Transformer, using only a fifth of its memory.}
\label{table:comp-baselines}
\begin{center}
\setlength{\tabcolsep}{1.0mm}{
\begin{tabular}{l|rrr|ccc|cc|c}
\multicolumn{1}{c|}{\bf MODEL}  & \multicolumn{3}{c|}{\bf MEMORY} & \multicolumn{3}{c|}{\bf PG19} & \multicolumn{2}{c|}{\bf arXiv} & {\bf C4(4K+)} \\
& \multicolumn{1}{c}{All} & \multicolumn{1}{c}{Short} & Long & Meena & T5 & Custom & Meena & Custom & Custom\\
\hline 




 & & & & & & & &\\[-2mm]
Transformer XL      & 13.6M & \scriptsize{13.6M} & \scriptsize{0M} & 8.65 & 11.41 & 12.42 & 2.60 & 3.22 & 18.22 \\
Block Recurrent      & 13.1M & \scriptsize{13.1M} & \scriptsize{0M} & 8.30  & 10.98 & 11.90 &  2.26 &  2.70 & 17.82 \\
\textbf{\melodi{} $S_{192}$+$L_{32}$}& {\bf11.0M} & \scriptsize{2.6M}& \scriptsize{8.4M} & {\bf8.08} & {\bf10.51} & {\bf11.47} & {\bf2.12} & {\bf2.54} & {\bf17.55}\\[2mm]

Memorizing Trans.   & 147.8M & \scriptsize{13.6M} & \scriptsize{134.2M} & 8.07 & 10.62 & 11.53 & 2.14 & 2.56 & 17.37 \\
\textbf{\melodi{} $S_{128}$+$L_{64}$} &  {\bf18.5M} & \scriptsize{1.7M} & \scriptsize{16.8M} & 8.06 & 10.44 & 11.42 & 2.11 & 2.52 & 17.53\\
\textbf{\melodi{}} $S_{192}$+$L_{96}$       & 27.8M & \scriptsize{2.6M} & \scriptsize{25.2M} & \textbf{7.91} & \textbf{10.29} & \textbf{11.27} & \textbf{2.09} & \textbf{2.49} & \textbf{17.25} \\
\end{tabular}
}
\end{center}
\vspace{-5mm}
\end{table}

\textbf{Comparisons:}
Table~\ref{table:comp-baselines} presents a comparison of \melodi{} against three baseline models (Transformer XL, Block Recurrent Transformer, and Memorizing Transformer) across three datasets. We evaluate three \melodi{} configurations: $S_{192}+L_{32}$, $S_{128}+L_{64}$, and $S_{192}+L_{96}$, where $S$ and $L$ denote the number of short-term and long-term tokens per context window, respectively.  All models (\melodi{} and baselines) utilize a 13-layer transformer architecture, except for Block Recurrent Transformer, which inserts a block recurrent layer into a 12-layer transformer, ensuring a similar parameter count for all models.

\melodi{} $S_{192}$+$L_{32}$ consistently outperforms both Transformer XL and Block Recurrent Transformer across all datasets while utilizing less memory. For instance, it achieves a perplexity of 10.51 on PG-19 (T5 vocabulary), surpassing Transformer XL (11.41) and Block Recurrent Transformer (10.98) while consuming only 11.0M memory compared to their 13.6M and 13.1M, respectively.

In comparison to the Memorizing Transformer, \melodi{} $S_{128}+L_{64}$ exhibits slightly improved performance while significantly reducing memory consumption by approximately 8 times. \melodi{} $S_{192}+L_{96}$ further improves perplexity across all datasets, achieving a substantial reduction in memory usage exceeding a factor of 5.
These trends remain consistent across different network depths (12-layer and 13-layer), as shown in Table~\ref{table:comp-baselines-more} in Appendix \ref{appendix:more-comparison}. These results collectively highlight \melodi{}'s efficacy and efficiency as a memory architecture for language models.



\subsection{Ablations}
We conduct an ablation study of \melodi{} using the default configuration ($S_{128}$+$L_{64}$) with 128 short-term and 64 long-term tokens per context window. The long-term memory spans 128 context windows, and the transformer architecture consists of 13 layers. All models are trained on the PG-19 dataset with the T5 vocabulary for 200k steps.

\begin{figure}[t]
\begin{center}
\centerline{\includegraphics[width=1.0\columnwidth]{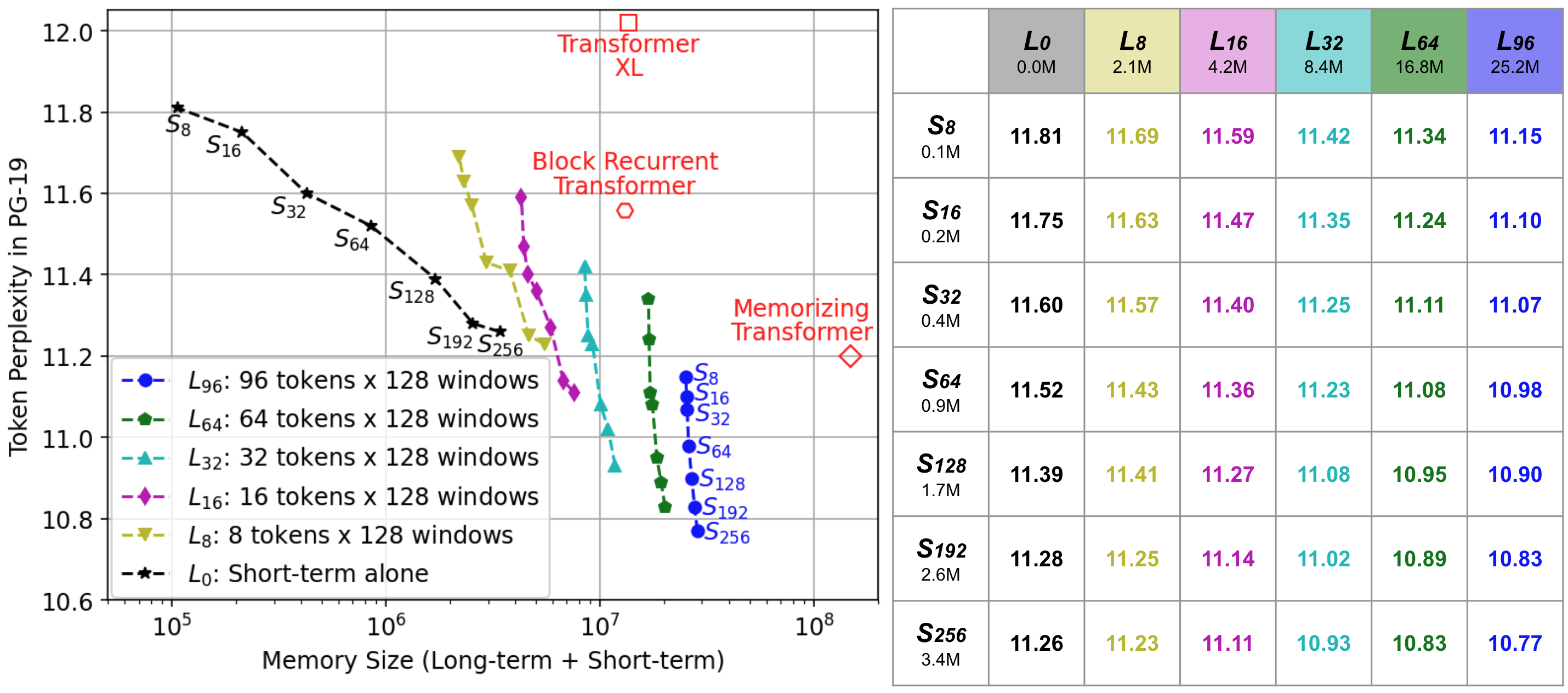}}
\caption{\textbf{Ablation of memory size on PG-19.} 
The token perplexity is reported for various combinations of short-term and long-term memory sizes. Each curve represents a fixed size of long-term memory, with points along the curve indicating different short-term memory sizes. For example, the \textcolor{blue}{blue curve} ($L_{96}$), uses 96 long-term tokens per window over 128 windows, totaling 12,288 tokens. Each point on this curve represents a different short-term memory size (e.g. $S_8$ denotes 8 short-term tokens per context window).
Memory size is measured by the number of floating-point numbers (floats). For instance, $L_{96}$ stores 12,288 long-term key-value (KV) pairs, each with 1024 dimensions, resulting in a total of 12,288$\times$1024$\times$2=25.2 million floats. The table on the right provides the perplexity results for each point on the left plot, using matching colors.
These results highlight that long-term and short-term memories play complementary roles, and increasing either type's capacity improves performance. Notably, \melodi{} achieves superior performance compared to baselines like Transformer XL, Block Recurrrent Transformer and Memorizing Transformer while utilizing fewer memory resources. Best viewed in color.
}
\label{fig:ablation-memory-size}
\end{center}
\vspace{-10mm}
\end{figure}

\textbf{Complementary roles of short-term and long-term memory:} 
Figure~\ref{fig:ablation-memory-size} illustrates how the sizes of both short-term and long-term memory jointly influence perplexity. Each curve represents a fixed long-term memory size, with varying short-term memory sizes depicted by points along the curve. 
With the exception of the black curve, which solely utilizes short-term memory, all other curves incorporate long-term memory spanning 128 context windows.
The figure demonstrates that increasing either short-term or long-term memory capacity leads to improved perplexity, highlighting their complementary roles in performance. Notably, by judiciously selecting memory sizes (e.g., $S_{192}$ for short-term and $L_{32}$ for long-term), we can outperform the Memorizing Transformer while utilizing less memory than Transformer XL and Block Recurrent Transformer.

\begin{table*}[t!]
\begin{minipage}[b]{0.325\linewidth}
        \begin{center}
		\includegraphics[width=1.0\linewidth]{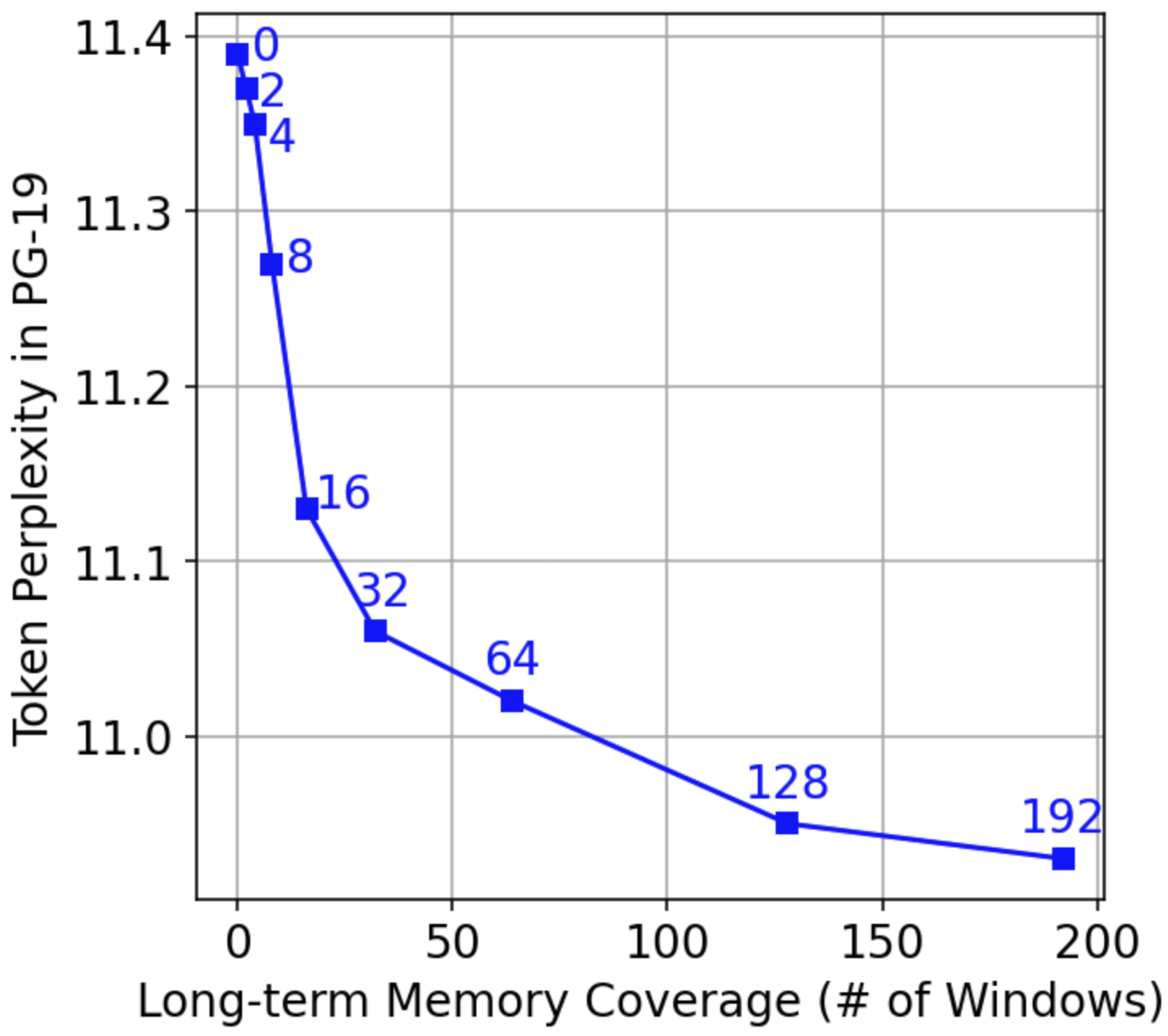}
	\end{center}
	\vspace{-4mm}
	\captionof{figure}{\textbf{Long-term memory coverage.} The coverage metric indicates the number of preceding context windows spanned by the long-term memory. All data points use $L$=64 long-term and $S$=128 short-term tokens per window. However, they vary in long-term memory capacity. For instance, `32' denotes covering 32 context windows, totaling 32$\times$64=2048 long-term tokens. 
	Perplexity improves marginally with long-term memory coverage of 2-4 windows, then accelerates until 32 windows, after which it levels off. 
	}
	\label{fig:long-term-ablation}
\end{minipage}
\quad
    \begin{minipage}[b]{0.645\linewidth}
	\begin{center}
		\includegraphics[width=1.0\linewidth]{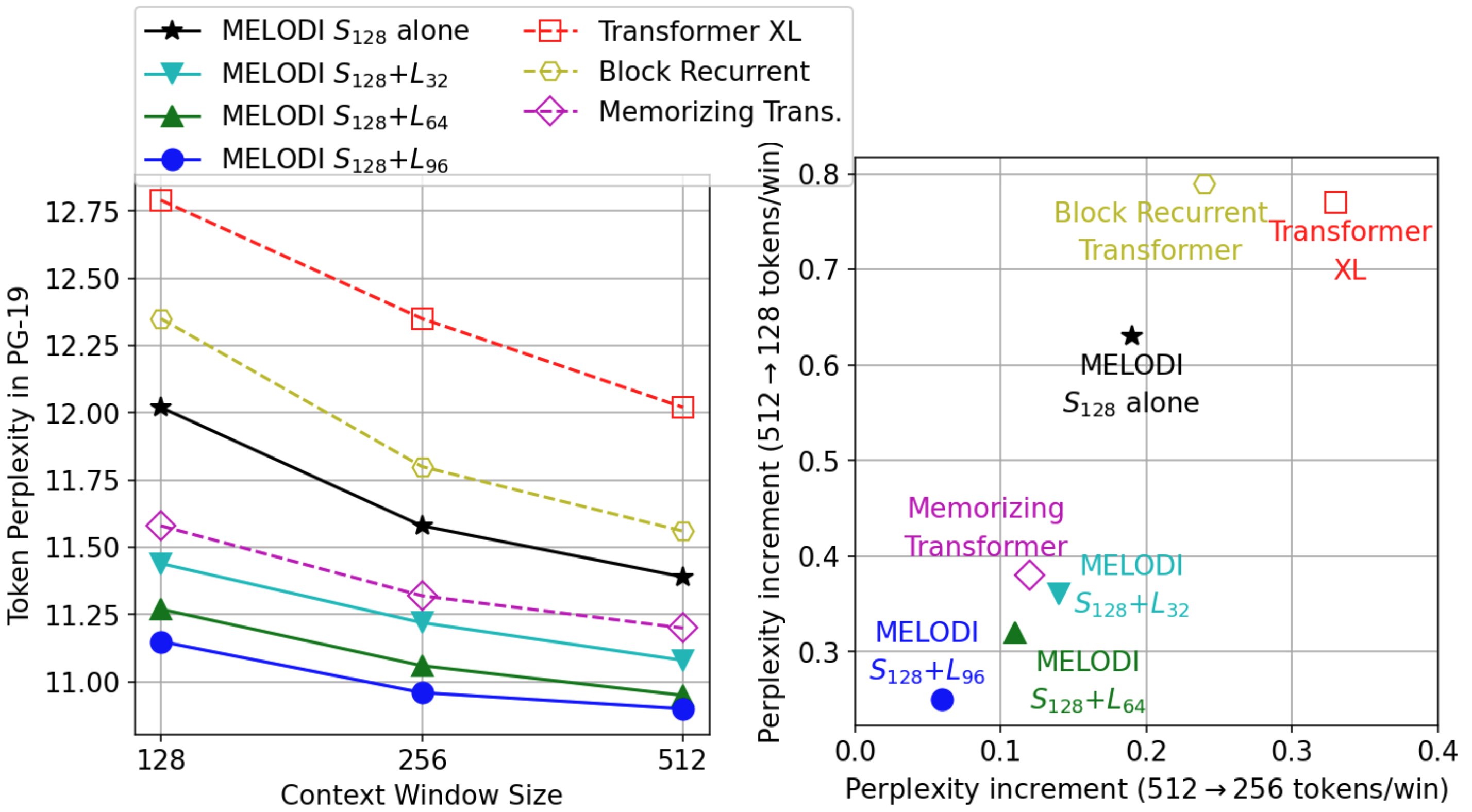}
	\end{center}
	\vspace{-4mm}
	\captionof{figure}{\textbf{Shorter context windows.} \textbf{(\textit{Left})}: \textit{Perplexity with varying context window sizes (512, 256, 128 tokens).} The number of short-term and long-term memory tokens in \melodi{} is proportionally adjusted ($\times\frac{1}{2}$, $\times\frac{1}{4}$, respectively) to ensure consistent long-term memory coverage. Even with smaller windows, \melodi{} with appropriate memory allocation (e.g. $S_{128}$+$L_{64}$) consistently outperforms baselines.
	\textbf{(\textit{Right}):} \textit{Perplexity increase due to window size reduction.} The $x$ and $y$ axes represent the perplexity increment when reducing the window size from 512 to 256 and 128 tokens, respectively. 
	 Models with long-term memory (i.e. \melodi{} variants with long-term memory and Memorizing Transformer) exhibit significantly less performance degradation (smaller perplexity increments) compared to those relying solely on short-term memory (\melodi{} with only short-term memory, Transformer XL, and Block Recurrent Transformer).}
	\label{fig:context-window}
    \end{minipage}  
    \vspace{-10mm}
\end{table*}

\textbf{Impact of long-term memory coverage:} 
In contrast to the previous ablation, we now maintain a fixed number of short-term ($S=128$) and long-term tokens ($L=64$) per context window while varying the number of windows encompassed by the long-term memory. Figure~\ref{fig:long-term-ablation} demonstrates that  performance improves as the long-term memory covers a wider range of context windows.

Interestingly, incorporating long-term memory for only the preceding 2 or 4 windows yields marginal perplexity improvements, suggesting that recent context is already effectively captured by the short-term memory. However,  performance gains accelerate as the long-term memory expands to encompass up to 32 windows, after which the improvements level off. This observation indicates that while the middle and distant history are beneficial for language modeling, they are not adequately retained in the short-term memory.  These findings further underscore the complementary nature of short-term and long-term memory mechanisms.

\textbf{Performance with shorter context windows:} 
%
%
Figure~\ref{fig:context-window} examines the impact of reducing context window size on model performance. The plot on the left displays perplexity scores for context window sizes of 512, 256, and 128 tokens. The number of short-term and long-term memory tokens in \melodi{} is proportionally adjusted ($\times\frac{1}{2}$, $\times\frac{1}{4}$, respectively) to ensure consistent long-term memory coverage across different window sizes.  Even with shorter context windows, \melodi{} with appropriate memory allocation (e.g., $S_{128}+L_{64}$) consistently outperforms the baseline models.

The plot on the right illustrates the increase in perplexity resulting from reducing the context window size. 
Notably, models incorporating long-term memory (\melodi{} variants with long-term memory and Memorizing Transformer) exhibit significantly less performance degradation (smaller perplexity increments) compared to models relying solely on short-term memory (\melodi{} variant with only short-term memory, Transformer XL, and Block Recurrent Transformer). This highlights the greater robustness of long-term memory mechanisms to reductions in context window size.

\textbf{Number of short-term layers:}
We investigate the effect of varying the number of short-term layers on model performance. For a given layer count (e.g., 4 layers), the short-term layers are uniformly distributed throughout the network (e.g., layers 1, 5, 9, and 13).  To disable short-term memory within a layer, we remove (a) the attention mechanism to the preceding short-term memory and (b) the linear token mixer responsible for updating the short-term memory.

Figure~\ref{fig:short-term-layers} shows that perplexity improves rapidly as the number of short-term layers increases from 1 to 4, after which the gains diminish. This observation supports our utilization of multiple layers for effective short-term memory modeling. However, it also suggests that disabling short-term memory in half of the layers offers a more efficient approach with negligible performance degradation.

\textbf{Number of long-term layers:}
Figure~\ref{fig:long-term-layers} illustrates the effect of varying the number of long-term layers on perplexity. We consider two configurations: $L_{64}$ and $L_{32}$, which compress each context window into 64 and 32 long-term tokens, respectively.
With $L_{64}$, introducing the first long-term layer yields a substantial perplexity improvement, after which performance plateaus.  In contrast, with $L_{32}$, while adding a second long-term layer provides a notable gain, the resulting performance remains inferior to that of a single $L_{64}$ layer, even though both configurations utilize the same total number of long-term tokens. These findings validate our design choice of employing a single long-term memory layer with a sufficient number of long-term tokens.

\textbf{Summary branching:}
Table~\ref{table:summary-branch} examines the effect of summary branching on perplexity, both with (ST+LT) and without (ST) long-term memory.  Summary branching consistently yields a perplexity improvement of approximately 0.3, indicating distinct summary information flow across network layers and context windows.

\begin{table*}[t]
\begin{minipage}[b]{0.29\linewidth}
    \caption{\textbf{Summary branching.} Summary branching provides a consistent gain of approximately 0.3 in perplexity, both with (column ST+LT) and without (column ST) long-term memory.}
    \begin{center}
    \setlength{\tabcolsep}{0.8mm}{
    \begin{tabular}{c|cc}
    {\bf Branching} & {\bf ST} & {\bf ST+LT} \\
    \hline 
    & & \\
    No &  11.68 &  11.24 \\
    Yes & {\bf 11.39} & {\bf 10.95} 
    \end{tabular}
    }
    \end{center}
    \label{table:summary-branch}
\end{minipage}
\quad
\begin{minipage}[b]{0.32\linewidth}
        \begin{center}
		\includegraphics[width=1.0\linewidth]{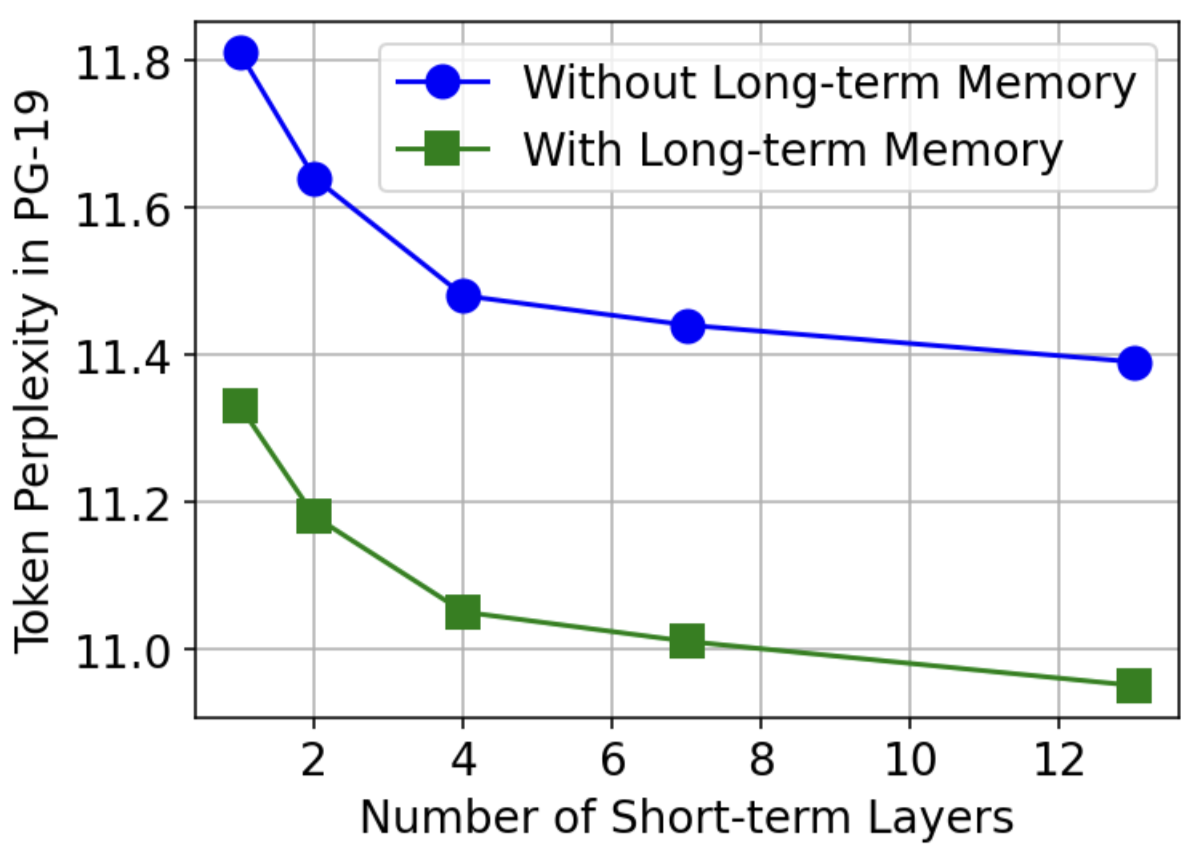}
	\end{center}
	\vspace{-4mm}
	\captionof{figure}{\textbf{Number of short-term layers.} Perplexity improves as the number of short-term layers increases.
	}
	\label{fig:short-term-layers}
\end{minipage}
\quad
\begin{minipage}[b]{0.32\linewidth}
	\begin{center}
		\includegraphics[width=1.0\linewidth]{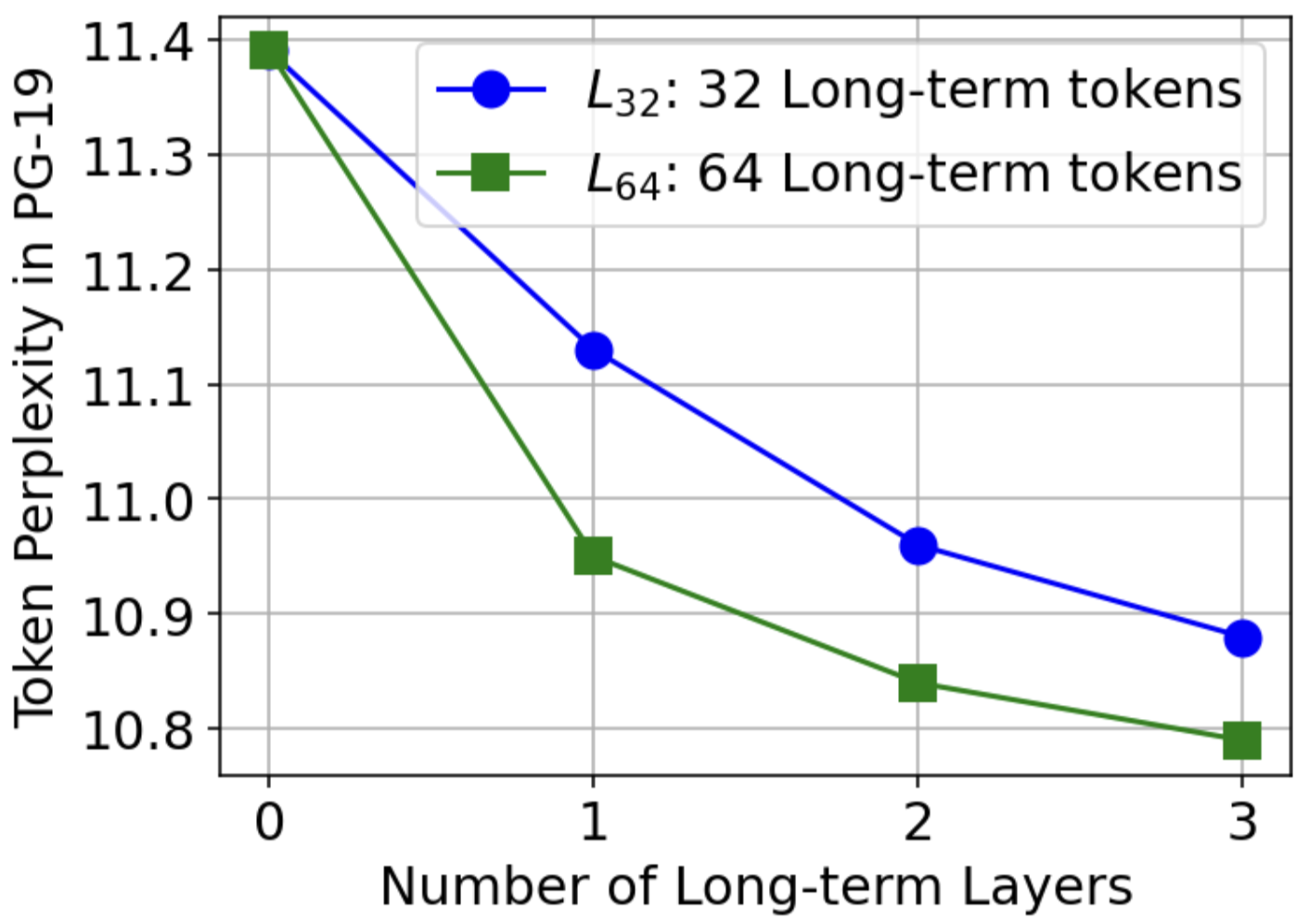}
	\end{center}
	\vspace{-4mm}
	\captionof{figure}{\textbf{Number of long-term layers.} A single layer with sufficient long-term tokens is adequate.}
	\label{fig:long-term-layers}
\end{minipage}  
\vspace{-2mm}
\end{table*}

\section{Related Work}
\textbf{Memory in language models:}
While LSTMs \citep{HochSchm97} achieve long-range memory through recurrent compression at the token level, the advent of Transformers \citep{NIPS2017_transformer} has shifted the focus to memory mechanisms operating at the context window level. Transformer-XL \citep{dai-etal-2019-transformer} introduces a caching mechanism to store key-value (KV) pairs from the preceding context window as a form of short-term memory. Block Recurrent Transformer \citep{NEURIPS2022_block_recurrent} and RMT \citep{NEURIPS2022_rmt}, integrate recurrent mechanisms inspired by LSTMs into Transformer architectures at the window level. \cite{munkhdalai2024leavecontextbehindefficient} explore the use of additional memory like Hopfield Networks \citep{hopfield-neural-networks-and-1982, ramsauer2021hopfieldnetworksneed}. Memorizing Transformer \citep{wu2022memorizing} introduces a dedicated memory layer to store KV pairs for long-term memory, whereas MemoryLLM \citep{icml2024memoryllm} incorporates long-term memory in every layer, incurring a substantial memory overhead. In contrast to these approaches, \melodi{} integrates both short-term and long-term memory into a transformer model via compression.

\textbf{Extending context length:}
Recent research has demonstrated promising progress in scaling the context length of language models. To mitigate the cost of attention mechanisms over long contexts, LongLoRA \citep{chen2024longloraefficientfinetuninglongcontext}, PCW \citep{ratner2023parallel}, and Landmark-Attention \citep{NEURIPS2023_landmark_attn} employ sparse local attention for efficient fine-tuning. Position-Interpolation \citep{chen2023extendingcontextwindowlarge} and YaRN \citep{peng2023yarnefficientcontextwindow} extend context size by modifying the RoPE embedding scheme \citep{su2023roformerenhancedtransformerrotary}.  \cite{xiong2023effectivelongcontextscalingfoundation} provide a practical recipe for extending LLAMA2 \citep{touvron2023llama2openfoundation} to handle up to 32,768 tokens, while \cite{fu2024dataengineeringscalinglanguage} further explore data-centric approaches for extending context length through lightweight continual pretraining.

\textbf{Compression:}
Recent work has demonstrated the effectiveness of Transformer models in compressing input sequences by appending summary tokens \citep{raecompressive2019, NEURIPS2022_rmt, chevalier-etal-2023-auto-compressor, ge2024incontextautoencodercontextcompression}. For instance, RMT \citep{NEURIPS2022_rmt} utilizes the output of summary tokens recurrently as short-term memory.  AutoCompressor \citep{chevalier-etal-2023-auto-compressor} aggregates summary tokens across segments to generate a summary representation for long documents used in retrieval tasks. GISTING \citep{mu2024learningcompresspromptsgist} applies this technique to compress long prompts.  ICAE \citep{ge2024incontextautoencodercontextcompression} further incorporates LoRA fine-tuning \citep{hu2021loralowrankadaptationlarge} for context compression, while TransformerFAM \citep{hwang2024transformerfamfeedbackattentionworking} introduces feedback attention to enhance performance.


\section{Conclusion}
In this work, we have introduced \melodi{}, a novel memory architecture designed to address the challenges of long document processing within the constraints of short context windows. The core innovation of \melodi{} lies in its hierarchical compression approach, wherein short-term memory facilitates smooth transitions between context windows through recurrent compression across multiple layers, and long-term memory preserves crucial information from the entire history by performing further compression and aggregation within a single middle layer. Our empirical evaluations on multiple long-context datasets have validated \melodi{} as an efficient and effective solution. The success of \melodi{} underscores the potential of hierarchical memory compression for tackling the complexities of long document processing. We anticipate that further research in this direction will enhance long context understanding and generation over multiple modalities.

\newpage

\bibliography{iclr2025_conference}
\bibliographystyle{iclr2025_conference}

\newpage

\appendix
\section{Limitations}
A primary limitation of our current method is its focus on training from scratch, without addressing the fine-tuning of pre-trained models with fixed context window sizes.  In future work, we plan to explore adapting \melodi{} (short-term and long-term memory) to enhance the long-context capabilities of pre-trained models through techniques like LoRA fine-tuning.

\section{Training Details}
\label{appendix:training-details}
We use Adafactor optimizer \citep{ICML-2018-ShazeerS} with a learning rate schedule that employs a linear warmup for the first 1000 steps, followed by cosine decay. The maximum and minimum learning rates are set to 0.01 and 0.001, respectively, as recommended in \cite{hoffmann2022trainingcomputeoptimallargelanguage}.  A dropout rate of 0.05 is applied. All models are trained for 500k steps (200k for ablations) on 32 TPU cores with a batch size of 32 (1 per core).

\begin{table}[b!]
\caption{\textbf{Comparisons with baselines on three datasets}. The table reports average token-level perplexities for various models on three datasets. All models were trained under the same settings (e.g.  segment length 4096, context window 512, 500k training steps). Three \melodi{} configurations were used: $S_{192}+L_{32}$, $S_{128}+L_{64}$, and $S_{192}+L_{96}$. For instance, $S_{192}+L_{32}$ indicates $S=192$ short-term and $L=32$ long-term tokens per context window. All \melodi{} models utilized a long-term memory spanning 128 context windows.
\melodi{} $S_{192}+L_{32}$ outperformed Transformer XL and Block Recurrent Transformer while consuming less memory. Notably, \melodi{} $S_{192}+L_{96}$ clearly surpassed Memorizing Transformer, using only a fifth of its memory.}
\label{table:comp-baselines-more}
\begin{center}
\setlength{\tabcolsep}{1.0mm}{
\begin{tabular}{l|rrr|ccc|cc|c}
{\bf MODEL}  & \multicolumn{3}{c|}{\bf MEMORY} & \multicolumn{3}{c|}{\bf PG19} & \multicolumn{2}{c|}{\bf arXiv} & {\bf C4(4K+)} \\
& \multicolumn{1}{c}{All} & \multicolumn{1}{c}{Short} & Long & Meena & T5 & Custom & Meena & Custom & Custom\\
\hline 

{\bf 12 LAYERS} & & & & & & & &\\[1mm]
Transformer XL      & 12.6M & \scriptsize{12.6M} & \scriptsize{0M} &  8.76 & 11.54 & 12.63 & 2.61 & 3.23 & 18.61 \\
Block Recurrent      & 12.1M & \scriptsize{12.1M} & \scriptsize{0M} & 8.47  & 11.12 &  12.11 & 2.27 & 2.73 & 18.27 \\
\textbf{\melodi{} $S_{192}$+$L_{32}$}& {\bf 10.8M} & \scriptsize{2.4M}& \scriptsize{8.4M} & {\bf8.22} & {\bf10.66} & {\bf11.66} & {\bf2.13 }& {\bf2.55} & {\bf18.03} \\ [2mm]

Memorizing Trans.   & 146.8M & \scriptsize{12.6M} & \scriptsize{134.2M} & 8.15 & 10.74 & 11.68 & 2.15 & 2.57 & 17.88 \\

\textbf{\melodi{} $S_{128}$+$L_{64}$} & {\bf18.4M} & \scriptsize{1.6M} & \scriptsize{16.8M} & 8.16 & 10.61& 11.66 & 2.13 & 2.55 & 18.01 \\
\textbf{\melodi{} $S_{192}$+$L_{96}$}& 27.6M &\scriptsize{2.4M} & \scriptsize{25.2M} & \textbf{8.08} & \textbf{10.48} & \textbf{11.47} & \textbf{2.11} & \textbf{2.51} & \textbf{17.75} \\[1mm]

\hline
{\bf 13 LAYERS} & & & & & & & &\\ [1mm]
Transformer XL      & 13.6M & \scriptsize{13.6M} & \scriptsize{0M} & 8.65 & 11.41 & 12.42 & 2.60 & 3.22 & 18.22 \\
Block Recurrent      & 13.1M & \scriptsize{13.1M} & \scriptsize{0M} & 8.30  & 10.98 & 11.90 &  2.26 &  2.70 & 17.82 \\
\textbf{\melodi{} $S_{192}$+$L_{32}$}& {\bf11.0M} & \scriptsize{2.6M}& \scriptsize{8.4M} & {\bf8.08} & {\bf10.51} & {\bf11.47} & {\bf2.12} & {\bf2.54} & {\bf17.55}\\[2mm]

Memorizing Trans.   & 147.8M & \scriptsize{13.6M} & \scriptsize{134.2M} & 8.07 & 10.62 & 11.53 & 2.14 & 2.56 & 17.37 \\
\textbf{\melodi{} $S_{128}$+$L_{64}$} &  {\bf18.5M} & \scriptsize{1.7M} & \scriptsize{16.8M} & 8.06 & 10.44 & 11.42 & 2.11 & 2.52 & 17.53\\
\textbf{\melodi{}} $S_{192}$+$L_{96}$       & 27.8M & \scriptsize{2.6M} & \scriptsize{25.2M} & \textbf{7.91} & \textbf{10.29} & \textbf{11.27} & \textbf{2.09} & \textbf{2.49} & \textbf{17.25} \\
\end{tabular}
}
\end{center}
\vspace{-5mm}
\end{table}

\section{More Comparison with Baselines}
\label{appendix:more-comparison}


\textbf{Comparison with baselines using a fixed context window size:}
Table~\ref{table:comp-baselines-more} presents a comparison of \melodi{} against three baseline models (Transformer XL, Block Recurrent Transformer, and Memorizing Transformer) across three datasets, using a consistent segment length of 4096 tokens and a context window size of 512. The evaluation includes both 12-layer and 13-layer transformer architectures to assess the impact of model depth on performance.

Notably, even with fewer layers, \melodi{} $S_{192}$+$L_{32}$ consistently outperforms both Transformer XL and Block Recurrent Transformer across all datasets while consuming less memory. For instance, the 12-layer variant of \melodi{} $S_{192}$+$L_{32}$ achieves a perplexity of 10.66 on PG-19 (T5 vocabulary), surpassing the 13-layer variants of Transformer XL (11.41) and Block Recurrent Transformer (10.98).

In comparison to the Memorizing Transformer, \melodi{} $S_{128}+L_{64}$ exhibits slightly improved performance while dramatically reducing memory consumption by a factor of 8.  Furthermore, \melodi{} $S_{192}+L_{96}$  achieves even better perplexity scores across all datasets with a substantial reduction in memory usage exceeding a factor of 5. The improvement is consistent for using both 12-layer and 13-layer transformer architectures. These results collectively highlight \melodi{}'s efficacy and efficiency as a memory architecture for language models.

\textbf{Comparison with baselines that utilize longer context windows:} 
Table~\ref{table:comp-baselines-longer} demonstrates that on PG-19 (T5 vocabulary), \melodi{} (with a context window size of 512 tokens) outperforms three baselines even when they utilize longer context windows (1024 and 2048 tokens), further highlighting its effectiveness and efficiency in processing long sequences. While the baselines generally exhibit improved performance with increasing context window size, Memorizing Transformer shows instability with a 2048-token window. We hypothesize that this instability stems from the limited backpropagation through time (BPTT) depth (two windows) within each segment (4096 tokens). 

\begin{table}[t]
\caption{\textbf{Comparisons with baselines that utilize longer context windows}. 
\melodi{}, with a context window size of 512, outperforms three baselines (i.e., Transformer XL, Block Recurrent Transformer and Memorizing Transformer) even when they utilize longer context windows (1024 and 2048 tokens). This highlights \melodi{}'s effectiveness in processing long sequences despite using shorter context windows.
}
\label{table:comp-baselines-longer}
\begin{center}
\setlength{\tabcolsep}{2.0mm}{
\begin{tabular}{l|rr|rrr|c}
\multicolumn{1}{c|}{\bf MODEL}  & {\bf Segment} & {\bf Window} & \multicolumn{3}{c|}{\bf MEMORY} & \multicolumn{1}{c}{\bf PG19} \\
& {\bf Length} & {\bf Length} & \multicolumn{1}{c}{All} & \multicolumn{1}{c}{Short} & Long & T5 \\
\hline 
 & & & & & & \\
Transformer XL & 4096 & 512 & 13.6M & \scriptsize{13.6M} & \scriptsize{0M} & 11.41 \\
Transformer XL & 4096 & 1024 & 27.3M & \scriptsize{27.3M} & \scriptsize{0M} & 11.11 \\
Transformer XL & 4096 & 2048 & 54.5M & \scriptsize{54.5M} & \scriptsize{0M} & 10.99 \\ 
Transformer XL & 6144 & 3072 & 81.8M & \scriptsize{81.8M} & \scriptsize{0M} & 10.96 \\ [2mm]
Block Recurrent & 4096 & 512     & 13.1M & \scriptsize{13.1M} & \scriptsize{0M} &  10.98  \\
Block Recurrent & 4096 & 1024    & 25.7M & \scriptsize{25.7M} & \scriptsize{0M} &  10.91  \\
Block Recurrent & 4096 & 2048    & 50.9M & \scriptsize{50.9M} & \scriptsize{0M} &  10.88  \\ [2mm]

Memorizing Trans. & 4096 & 512  & 147.8M & \scriptsize{13.6M} & \scriptsize{134.2M} & 10.62  \\
Memorizing Trans. & 4096 & 1024  & 161.5M & \scriptsize{27.3M} & \scriptsize{134.2M} & 10.53  \\
Memorizing Trans. & 4096 & 2048  & 188.7M & \scriptsize{54.5M} & \scriptsize{134.2M} & \small{not stable}  \\[2mm]

\textbf{\melodi{} $S_{192}$+$L_{32}$} & 4096 & 512 & {\bf11.0M} & \scriptsize{2.6M}& \scriptsize{8.4M} & 10.51\\
\textbf{\melodi{} $S_{128}$+$L_{64}$} & 4096 & 512 &  18.5M & \scriptsize{1.7M} & \scriptsize{16.8M} & 10.44\\
\textbf{\melodi{}} $S_{192}$+$L_{96}$  & 4096 & 512     & 27.8M & \scriptsize{2.6M} & \scriptsize{25.2M} &  \textbf{10.29}  \\
\end{tabular}
}
\end{center}
\end{table}

\begin{table*}[t!]
\begin{minipage}[t]{0.455\linewidth}
        \caption{\textbf{Directly copying short-term tokens to long-term memory:} Here, we force the short-term and long-term to share the same number of tokens (e.g. $S_{64}$+$L_{64}$), and examine the impact of directly using short-term tokens as long-term tokens, bypassing the linear token mixer. This approach results in performance degradation compared to generating distinct long-term tokens.}
\label{table:short-as-long}
\vspace{-2mm}
\begin{center}
\setlength{\tabcolsep}{1.2mm}{
\begin{tabular}{c|ccc}
\multicolumn{1}{c|}{\bf Copying}  &\multicolumn{1}{c}{$S_{96}$+$L_{96}$} & \multicolumn{1}{c}{$S_{64}$+$L_{64}$} &
\multicolumn{1}{c}{$S_{32}$+$L_{32}$}
\\ \hline \\
Yes & 11.00 & 11.36 & 11.42 \\
No  & {\bf 10.92} & {\bf 11.08} & {\bf 11.25} \\
\end{tabular}
}
\end{center}
\end{minipage}
\quad
    \begin{minipage}[t]{0.510\linewidth}
	\caption{\textbf{Position of long-term layer.} Here, we use a 13-layer \melodi{} model, where layers are indexed from 0 to 12.  The model uses 128 short-term and 64 long-term tokens per context window.
	While the default position of the long-term layer is at layer 8, placing it at different layers between 5 and 11 yields consistent perplexity scores. 
	}
\label{table:long-term-pos}
\vspace{-1mm}
\begin{center}
\setlength{\tabcolsep}{1.8mm}{
\begin{tabular}{ccccc}
\multicolumn{1}{c|}{\bf Layer}  & 5 & 6 & 7 & 8$^*$ \\
 \hline 
\multicolumn{1}{c|}{} & & & & \\
\multicolumn{1}{c|}{\bf Perplexity} & 11.00 & 11.01 & 11.03 & 10.95 \\
 \\

\multicolumn{1}{c|}{\bf Layer}  & 9 & 10 & 11 \\
\hline
\multicolumn{1}{c|}{}& & & & \\
\multicolumn{1}{c|}{\bf Perplexity} & 10.96 & 10.95 & 10.94 \\
\end{tabular}
}
\end{center}
    \end{minipage}  
    \vspace{-2mm}
\end{table*}

\section{More Ablations}
In this section, we present additional ablation studies.

\textbf{Directly copying short-term memory to long-term memory:} This ablation experiment, conducted at the long-term layer, explores directly copying short-term memory tokens as long-term tokens.  Instead of generating long-term tokens using the linear token mixer, this approach utilizes the short-term tokens present at the long-term layer and stores them directly in the long-term memory. Here, we force the short-term and long-term to share the same number of tokens ($S=L$). The results in Table~\ref{table:short-as-long} indicate that this direct copying method leads to a performance degradation.

\textbf{Effect of long-term layer position:}
We investigate the impact of varying the position of the long-term layer within a 13-layer \melodi{} model ($S_{128}$+$L_{64}$). Layers are indexed from 0 to 12, with the default long-term layer position at layer 8.  Results in Table~\ref{table:long-term-pos} reveal consistent perplexity scores when the long-term layer is positioned between layers 5 and 11.

\end{document}

%% file: math_commands.tex
\usepackage{amsmath,amsfonts,bm}

\def\eqref#1{equation~\ref{#1}}

\def\1{\bm{1}}

\DeclareMathAlphabet{\mathsfit}{\encodingdefault}{\sfdefault}{m}{sl}
\SetMathAlphabet{\mathsfit}{bold}{\encodingdefault}{\sfdefault}{bx}{n}

